%% file: main.tex
\documentclass{article} % For LaTeX2e
\usepackage{iclr2027_conference,times}

\input{math_commands.tex}

\usepackage{hyperref}
\usepackage{url}
\usepackage{graphicx}
\usepackage{wrapfig}
\usepackage{capt-of}

\title{From Routing Signals to Selective Review: \\Visual regrounding in MoE VLMs}

\author{%
  \parbox[t]{\dimexpr\textwidth-2\tabcolsep\relax}{%
    \centering\normalfont
    \textbf{Hongzhu Guo}\textsuperscript{1,2}%
    \thanks{Work done while at UCLA.}
    \quad
    \textbf{Mohsen Fayyaz}\textsuperscript{1}
    \quad
    \textbf{Nanyun Peng}\textsuperscript{1}
    \\[0.4em]
    \textsuperscript{1}University of California, Los Angeles
    \qquad
    \textsuperscript{2}Peking University
    \\[0.4em]
    {\small
      \texttt{hongzhuguo@ucla.edu}
      \qquad
      \texttt{mohsenfayyaz@cs.ucla.edu}
      \qquad
      \texttt{violetpeng@cs.ucla.edu}
    }%
  }%
}

\iclrfinalcopy % Uncomment for camera-ready version, but NOT for submission.
\begin{document}
\raggedbottom

\maketitle
\fancyhead{}

\input{section/0-abstract}
\input{section/1-introduction}
\input{section/2-related-work}
\input{section/3-method}
\input{section/4-experiments}
\input{section/6-conclusion}
\newpage
\section*{AI use statement}
We used large language models to assist with several aspects of this work. Specifically, they were used to help write experimental scripts, including linear-probing training scripts; search for and identify potentially relevant prior work; check the manuscript for logical inconsistencies and grammatical errors; and improve the clarity and fluency of the writing. We also used large language models to produce an initial draft of parts of the experimental results section. All AI-generated code was reviewed and tested by the authors, and all suggested references were independently verified. The authors reviewed, revised, and take full responsibility for all content in the final manuscript.

\bibliographystyle{iclr2027_conference}
\bibliography{references}

\appendix
\input{section/appendix}

\end{document}

%% file: math_commands.tex
\usepackage{amsmath,amsfonts,bm}

\def\eqref#1{equation~\ref{#1}}
\def\1{\bm{1}}

\DeclareMathAlphabet{\mathsfit}{\encodingdefault}{\sfdefault}{m}{sl}
\SetMathAlphabet{\mathsfit}{bold}{\encodingdefault}{\sfdefault}{bx}{n}

%% file: section/0-abstract.tex
\begin{abstract}
Vision-language models (VLMs) may accept false visual premises, answering questions about a target object's color, count, location, or state even when it is absent. We call this reliability-critical behavior a target-absence grounding failure. Existing visual-grounding detectors primarily rely on generated responses, hidden states, or uncertainty measures. We present the first framework to leverage internal routing decisions in Mixture-of-Experts (MoE) VLMs to detect target absence before generation and guide selective correction. We extract target-token routing probabilities from Qwen3-VL-30B-A3B-Instruct and Gemma-4-26B-A4B-it, train a separate L2-regularized linear detector for each model, and use its predictions to selectively invoke a target-aware review prompt. Using routing alone, the Qwen and Gemma detectors achieve ROC-AUCs of 0.9988 and 0.9956 on GQA-Inpaint and retain 0.8095 and 0.7781 on the external OBER dataset, respectively. The resulting routing-gated policy improves end-to-end accuracy on GQA-Inpaint and OBER by +22.25\% and +12.17\% for Qwen, and by +13.42\% and +1.39\% for Gemma, without modifying model weights. Further analysis shows that the signal is localized to the target-object token, emerges in early MoE layers, and is distributed across partially substitutable experts. Although cross-dataset threshold shifts require recalibration, false-positive review causes limited harm overall, suggesting that intervention risk can be controlled through joint selection of the detector threshold and review prompt. Overall, we show that routing probabilities alone preserve actionable information about visual perception, allowing computation already produced by an MoE VLM to support low-cost detection and selective visual regrounding.
\end{abstract}

%% file: section/1-introduction.tex
\begin{figure}[htbp]
    \centering
    \includegraphics[width=\textwidth]{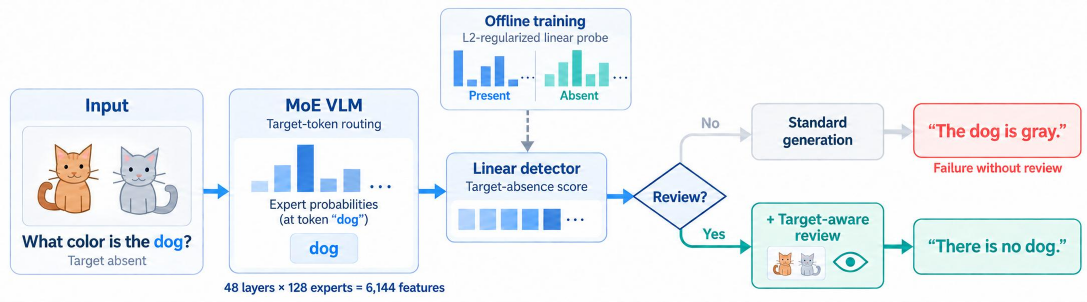}
    \caption{Overview of routing-gated visual regrounding. A linear detector is
trained offline on target-token routing probabilities collected
across MoE layers and experts. At inference time, the detector scores
routing from the standard-prompt prefill before answer generation.
Predicted-present inputs follow standard decoding, while
predicted-absent inputs trigger a target-aware prompt that asks the
unchanged VLM to verify the queried object's presence and explicitly
report its absence when unsupported.}
    \label{fig:overview}
\end{figure}

\section{Introduction}
\label{sec:introduction}
Vision-language models (VLMs) perform strongly on visual question answering and multimodal reasoning, yet their answers are not always grounded in the image. A consequential failure arises when a question contains a false visual premise. Given an image with no dog, for example, a model may answer ``What color is the dog?'' with a plausible color. Here, the object is supplied by the question, and the model accepts its premise and fabricates the requested attribute. We call this behavior a \emph{target-absence grounding failure}. Such answers can appear fluent and task-responsive because the prompt requests an attribute rather than an explicit existence judgment, obscuring the missing visual evidence and motivating detection before a response is produced.

Prior work evaluates object hallucination and false-premise questions through absent-object and object-removal benchmarks~\citep{li2023pope,lovenia2024nope,he2025removedobjects}. Rechecking the image, using an external verifier, or consulting a stronger model may mitigate such errors, but doing so for every input adds inference cost. Existing detectors primarily use generated responses, hidden states, or uncertainty measures~\citep{li2024referencefree,kogilathota2026halp}. Mixture-of-experts VLMs (MoE VLMs) expose a different signal: at every layer, a router already assigns each token to a subset of experts, providing a structured record of how computation is allocated and a direct opportunity to test whether visual support is reflected in that allocation~\citep{shazeer2017moe,fedus2022switch}. Whether this MoE-specific routing signal reveals target absence before generation, where it is located, and whether it can support selective correction remain unclear.

We study these questions using Qwen3-VL-30B-A3B-Instruct as the primary model and Gemma-4-26B-A4B-it to evaluate cross-model generalization. Our detect-then-review framework identifies the target noun phrase, maps it to tokenizer positions, and extracts routing probabilities specifically at that span across layers. A model-specific L2-regularized linear probe predicts target absence from this representation. Its prediction serves as a gate: predicted-present inputs retain the standard response, whereas predicted-absent inputs receive a target-aware review request. No model weights or router parameters are changed.

On GQA-Inpaint, the Qwen and Gemma routing probes reach test ROC-AUCs of 0.9988 and 0.9956, respectively. Their ranking signal transfers to the external OBER dataset, where the corresponding ROC-AUCs are 0.8095 and 0.7781, although a threshold calibrated on GQA transfers less reliably. When used to gate target-aware review, the method improves end-to-end accuracy for both models and on both datasets; for Qwen, the gains are 22.25 percentage points on GQA-Inpaint and 12.17 points on OBER. Further analysis shows that the signal emerges in early MoE layers and is distributed across many experts. False-positive review causes limited harm overall, although cross-dataset threshold shifts show that the ranking signal transfers more reliably than a fixed decision threshold.

These results reveal a mismatch between internal routing and final behavior: routing can detect missing visual evidence that the answer ultimately ignores. Routing can thus support both expert interpretation and low-cost monitoring of visual grounding, turning an interpretability observation into an actionable inference policy without an external vision model or repeated decoding. Selective review concentrates additional inference on likely failures rather than every request. Consequently, the extra review cost is incurred only on the triggered subset, reducing the expected inference overhead relative to reviewing every input.

Our contributions include: (a) demonstrating that target-object presence and absence are encoded in target-token routing, with the signal transferring across datasets and exhibiting identifiable layer- and expert-level structure; (b) training model-specific linear probes over layer--expert routing probabilities that identify target-absent inputs with high in-domain accuracy and generalize across two MoE VLM families; and (c) developing a routing-gated review method that improves end-to-end accuracy without modifying model weights, while quantifying the correction gains and false-positive risks of selective intervention.

%% file: section/2-related-work.tex
\section{Related Work}
\label{sec:related-work}

\subsection{Visual Grounding Failures and False-Premise Evaluation}
\label{sec:rw-hallucination}

Visual grounding evaluation asks whether a model's claims are supported by the image. Early work measured unsupported object mentions in captions through CHAIR; POPE subsequently made object presence the explicit subject of binary questions~\citep{rohrbach2018chair,li2023pope}. AMBER extends this evaluation to attributes and relations, recognizing that a response can be visually unsupported even beyond an incorrect existence judgment~\citep{wang2023amber}.

False-premise questions expose a different demand: rather than explicitly judging existence, the model must recognize that the requested information presupposes an absent object. NOPE evaluates questions about absent objects, requiring models to acknowledge missing visual evidence instead of supplying a plausible answer~\citep{lovenia2024nope}. Removed-object benchmarks make such negatives more challenging by deleting an object that originally belonged to the scene, preserving contextual cues that can still make its presence plausible~\citep{he2025removedobjects}. We use this setting to study detection and review of target-absence grounding failures, rather than treating every form of visual hallucination as the same prediction task.

\subsection{Internal Signals for Grounding Detection}
\label{sec:rw-internal-detection}

A central distinction among hallucination detectors is when sufficient evidence becomes available to assess an answer. Response-based approaches use confidence or consistency after producing candidate answers; INSIDE, for example, compares generated responses in representation space~\citep{li2024referencefree,chen2024inside}. Semantic Entropy Probes estimate semantic uncertainty from
hidden states without requiring multiple sampled responses
at inference time, and support both post-generation and
pre-generation probing \citep{kossen2024semanticentropyprobes}. Monitoring can also move inside that process: entity-level linear probes identify hallucinated tokens as a long-form response unfolds, without waiting for the full answer~\citep{obeso2026realtime}.

Other pre-generation probing methods estimate hallucination
risk from query representations before decoding begins
\citep{ji2024queryrisk}. HALP applies this principle to VLMs by examining visual features, decoder vision-token states, and multimodally integrated query states during prefill~\citep{kogilathota2026halp}. This establishes that early detection is possible, but leaves open which other internal representations can expose the visual support of a particular referent. We examine target-token MoE routing as such a representation, retaining explicit layer--expert coordinates for interpreting the detector's predictive signal.

\subsection{MoE Routing Analysis and Behavioral Control}
\label{sec:rw-moe}

Because MoE routers allocate tokens to selected experts, their decisions offer a direct view of how computation is organized~\citep{shazeer2017moe,fedus2022switch}. The organization is not uniform across models: Mixtral reports limited domain specialization, whereas OLMoE identifies clearer domain- and vocabulary-associated routing patterns~\citep{jiang2024mixtral,muennighoff2025olmoe}. Within models, expert diversity and sharing also vary with depth, as shown by cross-model analyses and the middle-layer cross-lingual alignment observed in Multilingual Routing~\citep{lo2025closer,bandarkar2026multilingualrouting}. These differences motivate studying routing under specific inputs and layers, rather than assigning a universal functional label to an expert.

Once routing patterns are associated with a behavior, they can also become targets for intervention. SteerMoE and RICE exploit this connection through inference-time expert steering, while Router Lens identifies context-faithful experts for selective fine-tuning~\citep{fayyaz2026steermoe,wang2025rice,bai2025routerlens}. These approaches use expert analysis to alter the model's computation or parameters. Our use is instead observational: unchanged routing supplies features for an absence detector, whose output selects a review prompt rather than directly steering experts.

\subsection{Visual Verification and Selective Review}
\label{sec:rw-selective-review}

Detecting unsupported visual premises is useful only if the resulting signal can guide an appropriate response. Existing mitigation methods provide several correction mechanisms: VCD and OPERA modify decoding, whereas Woodpecker verifies visual claims through an additional correction pipeline~\citep{leng2024vcd,huang2024opera,yin2024woodpecker}. VDGD takes another route, generating an image description and using it to ground subsequent answer decoding~\citep{ghosh2025vdgd}. These methods address how to correct, but the availability of a correction mechanism does not establish that it should be applied to every input.

Prompted reconsideration can introduce errors as well as resolve them, making the choice of when to intervene part of the reliability problem~\citep{kamoi2024selfcorrection}. RSP explicitly addresses this choice by thresholding pre-generation attention entropy or first-token confidence to trigger verification, without training a detector~\citep{huang2026rsp}. Its input-dependent benefits and harms motivate selective review, while leaving room for task-specific signals beyond general uncertainty~\citep{huang2026rsp}. We instantiate this selection with a supervised target-absence probe over MoE routing, linking its prediction to target-aware review and evaluating the resulting correction--harm trade-off.

%% file: section/3-method.tex
\section{Methodology}
\label{sec:method}

We consider a single-image detection and correction setting. Each example consists of an image $I_i$, a question $q_i$ referring to a target object, and a binary label $y_i\in\{0,1\}$, where $y_i=1$ indicates that the target is absent and $y_i=0$ indicates that the target is present. Controlled target-present and target-absent images may originate from the same source scene, but every $(I_i,q_i)$ is processed independently. The method follows the same order as inference. We first extract routing at the target-object tokens, then use it to estimate target absence, and finally invoke visual review only when that estimate exceeds a threshold.

\subsection{Target-Token Routing Representation}
\label{sec:target-routing}

We first identify the target noun phrase $\phi_i$ in $q_i$ with a frozen rule-based extractor, abstaining on unsupported or ambiguous questions. The VLM tokenizer then maps $\phi_i$ to its sub-token positions $T_i$ in the complete multimodal sequence; no gold object annotation is used at inference time.

During standard prompt prefill, we record the router's complete pre-selection distribution at these positions, average it over the target span, and concatenate the result across layers:
\begin{equation}
\mathbf{r}_{i,l,t}
=\operatorname{softmax}\!\left(W_l^{r}\mathbf{h}_{i,l,t}\right),\quad
p_{i,l,e}=\frac{1}{|T_i|}\sum_{t\in T_i}r_{i,l,t,e},\quad
\mathbf{p}_i=\operatorname{vec}\!\left([p_{i,l,e}]_{l,e}\right)
\in\mathbb{R}^{LE}.
\label{eq:routing-representation}
\end{equation}
Although only the Top-$k$ experts are executed, $\mathbf{p}_i$ retains probabilities for all $E$ experts. For Qwen3-VL-30B-A3B-Instruct, $L=48$, $E=128$, and $k=8$, so each input yields a 6,144-dimensional routing vector.

\subsection{Routing-Based Target-Absence Detection}
\label{sec:routing-detector}

Given $\mathbf{p}_i$, we next learn a detector of whether the target is absent. Let $j=(l,e)$ index one flattened layer--expert dimension. To prevent information leakage, we standardize it using only training-split statistics:
\begin{equation}
z_{i,j}=
\frac{p_{i,j}-\mu_j^{\mathrm{train}}}
{\max(\sigma_j^{\mathrm{train}},\epsilon)},
\label{eq:standardization}
\end{equation}
where $\epsilon$ is a small numerical constant. The same frozen statistics are applied to validation, test, and external data.

The standardized vector is passed to an L2-regularized linear probe, which produces an absence score
\begin{equation}
s_i=P(y_i=1\mid\mathbf{z}_i)
=\sigma\!\left(\mathbf{w}^{\top}\mathbf{z}_i+b\right).
\label{eq:linear-probe}
\end{equation}
We train the probe using binary cross-entropy with L2 regularization. We select the regularization strength by validation ROC-AUC and then select a decision threshold $\theta$ by validation balanced accuracy. The training statistics, probe parameters, and threshold are frozen before test and cross-dataset evaluation; test labels influence none of these choices (Appendix~\ref{app:implementation}, Table~\ref{tab:probe-training-configuration}).

Besides producing the score used for review, the linear form makes the detector directly interpretable at the layer--expert level. Across $S$ fitted training seeds, we rank dimension $j$ by the magnitude of its mean standardized coefficient:
\begin{equation}
\bar{w}_j=\frac{1}{S}\sum_{s=1}^{S}w_j^{(s)},
\qquad
R_j=|\bar{w}_j|.
\label{eq:expert-ranking}
\end{equation}
The sign preserves direction: $\bar{w}_j>0$ associates greater routing probability with target absence, whereas $\bar{w}_j<0$ associates it with target presence. Because expert probabilities within a layer are compositional and correlated, $R_j$ is a conditional predictive contribution rather than a standalone activation effect or causal importance. We therefore interpret this ranking together with the token-, layer-, and expert-level analyses in Section~\ref{sec:experiments}.

\subsection{Routing-Gated Selective Review}
\label{sec:selective-review}

Finally, we convert the absence score into an inference-time action. The frozen threshold defines a review trigger
\begin{equation}
g_i=\mathbf{1}\{s_i\geq\theta\}.
\label{eq:review-trigger}
\end{equation}
Let $G(I,q)$ denote generation by the unchanged MoE VLM, and let $\rho(q,\phi)$ augment the original question with a review instruction about target $\phi$. The routing-gated output is
\begin{equation}
o_i^{\pi}=
\begin{cases}
G(I_i,q_i), & g_i=0,\\
G(I_i,\rho(q_i,\phi_i)), & g_i=1.
\end{cases}
\label{eq:gating-policy}
\end{equation}
Thus, a predicted-present input keeps the standard response, whereas a predicted-absent input is regenerated from the same image and question with an added review instruction. We instantiate $\rho$ in two ways. The \emph{generic} instruction asks the model to inspect the image carefully; the \emph{target-aware} instruction instead asks it to verify whether $\phi_i$ is present and to state explicitly when it is absent (Appendix~\ref{app:implementation}, Table~\ref{tab:review-prompt-templates}).

This construction separates detection from correction: the probe decides when to intervene, while the review instruction specifies which visual premise the model should reconsider.

In an online implementation, $s_i$ is computed from routing already produced during the standard prompt prefill. If $g_i=0$, decoding proceeds along the standard path. If $g_i=1$, the reviewed prompt incurs one additional forward pass. The additional review cost therefore scales with the trigger rate rather than the full input stream. Neither branch modifies the model weights or router parameters; routing is used only to choose the appropriate prompting path.

%% file: section/4-experiments.tex
\section{Experiments}
\label{sec:experiments}

\subsection{Experimental Setup}
\label{sec:experimental-setup}

\textbf{Model and data.}
We train a separate routing probe for each of two MoE VLMs, Qwen3-VL-30B-A3B-Instruct and Gemma-4-26B-A4B-it, using 20,000 training pairs, 5,446 validation pairs, and 5,776 test pairs. All pairs pass automatic quality control, and images from the same source scene stay in the same split. Each pair contains one target-present image and one target-absent image. For the review experiment, we use 120 manually audited pairs from GQA and 115 manually audited pairs from OBER (filtering 5 with ambiguous semantics). Target presence in the original image and the absence of all same-category objects in the edited image are manually checked. The appendix gives the full data and annotation details.

\textbf{Review process.}
We test three prompts: standard, generic review, and target-aware review. Generic review asks the model to inspect the image before answering, and target-aware review asks it to check whether the target object is present. We sample five answers for each image--question pair under each prompt. Review is triggered when the detector score exceeds the model's calibrated threshold. We select each threshold on GQA validation and keep the standard answer when review is not triggered. The appendix gives the prompts, generation settings, training details, and thresholds.

\textbf{Evaluation protocol.}
Generated answers, including the added always-review answers, are manually checked against the images; Appendix~\ref{app:data-evaluation} gives the scoring rules. We report ROC-AUC for detector ranking and accuracy at the chosen threshold. ROC-AUC measures how well the detector ranks absent images above present images, and accuracy is the fraction of correctly classified images. Absence recall is the fraction of absent images sent to review, and specificity is the fraction of present images that do not trigger review. Trigger rate is the fraction of all images sent to review. Policy accuracy is the fraction of correct final answers across both present and absent images, so it counts both corrections and errors caused by review.

\subsection{Experiment Results}
\label{sec:main-results}

\subsubsection{Routing Gate Reliability}
\begin{figure*}[htbp]
    \centering
    \includegraphics[width=0.85\textwidth]{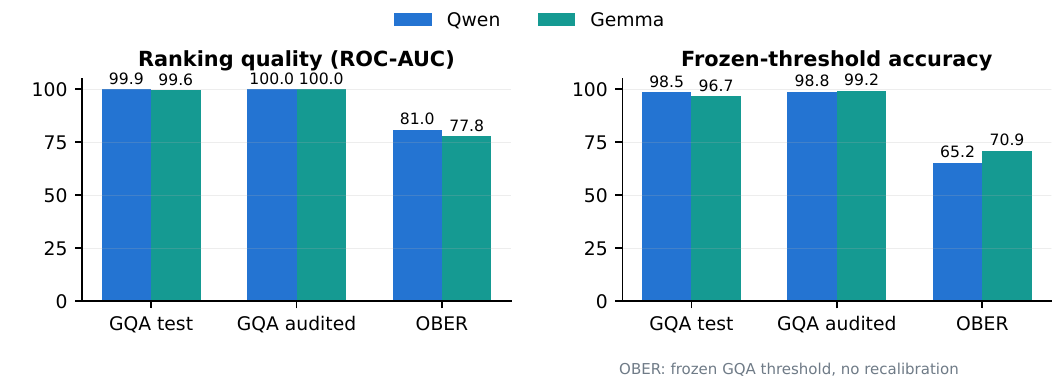}
        \caption{Routing-detector ROC-AUC and accuracy across models and datasets. Each model uses its GQA validation threshold on both datasets. The manually audited GQA set includes validation images and is used to test the review policy. Detection remains informative across datasets, with thresholds requiring recalibration for transfer.}
        \label{fig:detector-reliability}
\end{figure*}

Reliable detection is the first step in deciding which inputs need review. On the frozen GQA-Inpaint test split, Qwen and Gemma achieve 0.9988 and 0.9956 ROC-AUC (98.54\% and 96.68\% accuracy; Fig.~\ref{fig:detector-reliability}). Thus, target-token routing can predict target absence before the model answers.

As a control, we repeat Gemma probe training 20 times with separately shuffled training and validation labels. We select probes using the shuffled validation labels, then evaluate them against the true validation labels. Their ROC-AUC is $0.515\pm0.062$, close to chance.

\begin{wraptable}{r}{0.48\textwidth}
\vspace{-0.28in}
\centering
\footnotesize
\setlength{\tabcolsep}{4pt}
\caption{Qwen detection on the GQA-Inpaint held-out test split. All probes use validation-selected regularization and thresholds.}
\label{tab:representation-baselines}
\resizebox{\linewidth}{!}{%
\begin{tabular}{lrr}
\hline
Representation & ROC-AUC & Accuracy \\
\hline
Input embedding & 0.5000 & 50.00\% \\
Mean of hidden layers & 0.9979 & 98.06\% \\
Last hidden layer (L47) & 0.9980 & 98.16\% \\
All-layer routing & \textbf{0.9988} & \textbf{98.54\%} \\
\hline
\end{tabular}%
}
\end{wraptable}

Both detectors perform well on the manually audited GQA images and retain a useful ranking signal on OBER. However, the thresholds selected on GQA work less well on OBER. Qwen detects most absent targets but also sends many present images to review, while Gemma has a smaller gap between absence recall and specificity. Figure~\ref{fig:detector-reliability} shows the difference between ranking quality and accuracy at the selected threshold.

\begin{wrapfigure}{r}{0.42\columnwidth}
    \centering
    \includegraphics[width=0.8\linewidth]
        {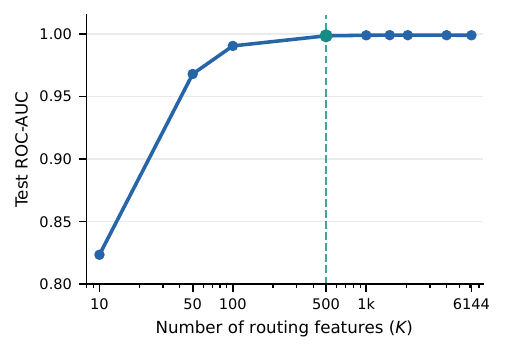}
    \caption{Test ROC-AUC on GQA-Inpaint using the top
        $K$ routing features. Performance nearly
        plateaus at $K=500$.}
    \label{fig:topk-probes}
\end{wrapfigure}

We compare routing with three target-token representation baselines: input embeddings, the mean hidden state across layers, and the final-layer hidden state. The first averages the target tokens' text embeddings before decoder processing, so it represents the target phrase without image context. The second averages the target-token hidden states across all decoder layers, capturing the representation after the model processes the image and question. The final-layer baseline uses the target-token hidden state from Layer 47 without averaging across layers. Each baseline yields 2,048 features, compared with the 6,144 layer--expert probabilities used by our routing probe. We train an L2-regularized linear probe for each representation using the same GQA-Inpaint examples, splits, and target-token spans, and select regularization and thresholds on validation data.

Table~\ref{tab:representation-baselines} shows that input embeddings give chance-level detection, whereas mean hidden states reach 0.9979 ROC-AUC and 98.06\% accuracy. All-layer routing performs slightly better, reaching 0.9988 ROC-AUC and 98.54\% accuracy. Routing thus offers a small performance gain over the mean-hidden-state baseline while retaining explicit layer--expert coordinates. These coordinates let us identify which expert probabilities contribute to the probe's prediction and examine their distribution across layers in Section~\ref{sec:analysis-experiments}. 

Since expert routing probabilities contribute unevenly to detection, we retrain probes on the top \(K\in\{10,50,100,500,1000,1500,2048,4000,6144\}\) routing features: eight reduced settings and the full-feature reference. Figure~\ref{fig:topk-probes} shows that ROC-AUC largely plateaus by \(K=500\) (0.9984 versus 0.9988 at \(K=6144\)). Thus, 500 features retain nearly all of the full probe's detection performance on GQA; this comparison does not measure VLM inference cost. Appendix~\ref{app:topk-probes} gives the feature-selection and training details.

\subsubsection{Routing-Gated Visual Regrounding}
\begin{figure*}[htbp]
    \centering
    \includegraphics[width=0.87\textwidth]{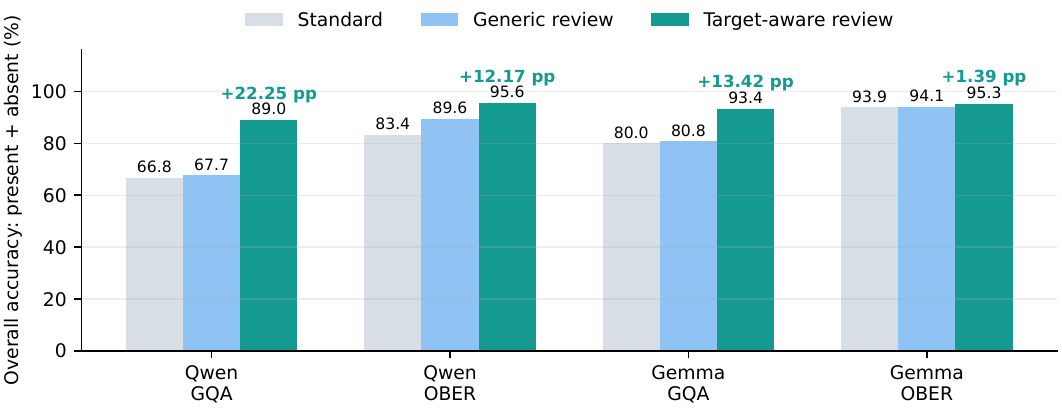}
        \caption{End-to-end accuracy of standard answering and routing-gated review. Accuracy is computed jointly over target-present and target-absent inputs. Target-aware review yields positive gains for both models and both datasets, while generic review is consistently weaker.}
        \label{fig:gated-review}
\end{figure*}

The routing gate selects inputs for review. Figure~\ref{fig:gated-review} shows that target-aware review raises Qwen accuracy from 66.75\% to 89.00\% on GQA and from 83.39\% to 95.57\% on OBER. Gemma improves from 80.00\% to 93.42\% and from 93.91\% to 95.30\%, respectively. Qwen's gains are positive in all five runs ($22.25\pm0.23$ and $12.17\pm0.81$ points). Generic review asks only for another look at the image; as a control, it gives smaller gains in all four settings (Appendix~\ref{app:policy-details}).

\begin{wraptable}{r}{0.48\textwidth}
\vspace{-0.29in}
\centering
\footnotesize
\setlength{\tabcolsep}{4pt}
\caption{Qwen target-aware accuracy by true target presence. Always review replaces every answer; selective review retains the standard answer when the gate does not trigger.}
\label{tab:selective-always-review}
\resizebox{\linewidth}{!}{%
\begin{tabular}{llrrr}
\hline
Dataset & Target & Standard & Always & Selective \\
\hline
GQA & Absent & 37.17\% & 83.17\% & 82.50\% \\
GQA & Present & 96.33\% & 89.33\% & 95.50\% \\
GQA & Total & 66.75\% & 86.25\% & \textbf{89.00\%} \\
\hline
OBER & Absent & 66.78\% & 95.83\% & 95.83\% \\
OBER & Present & 100.00\% & 94.78\% & 95.30\% \\
OBER & Total & 83.39\% & 95.30\% & \textbf{95.57\%} \\
\hline
\end{tabular}%
}
\end{wraptable}

Table~\ref{tab:selective-always-review} separates Qwen's results by true target presence. Target-aware review produces its gains on absent images, while always reviewing present images lowers their accuracy. The gate preserves most correct present-image answers. Across both groups, selective review outperforms always review on GQA (89.00\% versus 86.25\%) and OBER (95.57\% versus 95.30\%), while avoiding 49.58\% and 16.09\% of review calls, respectively. The OBER accuracy difference is small. Appendix~\ref{app:policy-details} reports the generic-control results (Tables~\ref{tab:presence-strata-full} and~\ref{tab:selective-always-overall}).

\subsubsection{Correction on Triggered Absent Images}
To isolate correction after a trigger, we also evaluate target-absent images sent to review. Target-aware review raises Qwen's OBER accuracy from 66.49\% to 95.79\%. Gemma starts at 93.10\% on this subset and reaches 100\%; its smaller gain reflects the limited room for improvement. Appendix Figure~\ref{fig:triggered-correction} gives all four model--dataset comparisons, including GQA.

\subsubsection{Cost of False Triggers}
To assess the cost of an incorrect review decision, we examine target-present images that the detector classifies as absent. We measure harm as the fraction of initially correct answers on these images that become wrong after review.

\begin{figure*}[htbp]
    \centering
    \includegraphics[width=0.9\textwidth]{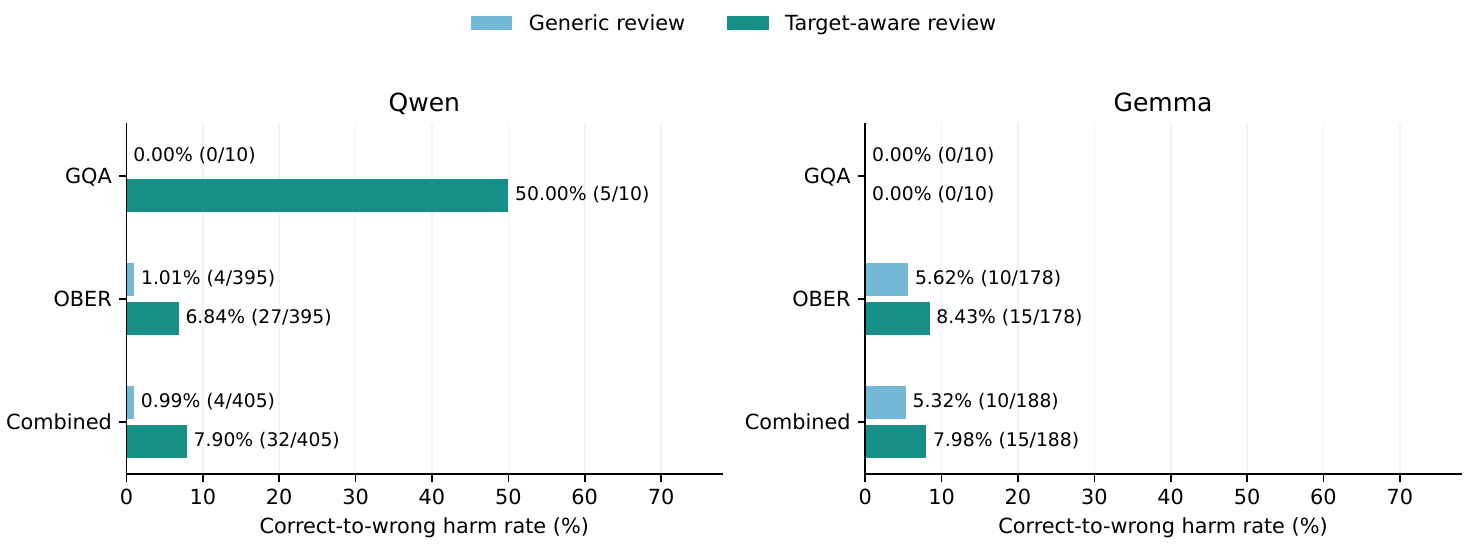}
    \caption{Correct-to-wrong harm after false review triggers. Bars show the percentage of initially correct answers on target-present images that become wrong after review; labels give the numerator and denominator. Qwen has 2 false-triggered images in GQA and 79 in OBER; Gemma has 2 and 36, respectively. Each image has five sampled answers.}
    \label{fig:false-trigger-harm}
\end{figure*}

Figure~\ref{fig:false-trigger-harm} shows that, across both datasets, generic review changes 4 of 405 initially correct Qwen answers to wrong answers (0.99\%), while target-aware review changes 32 (7.90\%). Gemma shows the same pattern, with 10 of 188 answers becoming wrong under generic review (5.32\%) and 15 under target-aware review (7.98\%). The GQA results also differ between models: target-aware review harms 5 of 10 initially correct Qwen answers but none of the 10 Gemma answers. However, each GQA subset contains only two images with five sampled answers per image, so these rates provide limited evidence about dataset-level risk.

Together with the correction gains in Appendix Figure~\ref{fig:triggered-correction}, these results show a trade-off: target-aware review corrects more absent-object answers, but causes more harm when present objects trigger review.

\subsection{Analysis Experiments}
\label{sec:analysis-experiments}

The main results show that routing can detect target absence and guide review. The following analyses examine how this information is used by the probe and where it can be read from the model. We first examine how predictive weight is distributed across experts, and then test whether the early-layer pattern identified by the full probe is independently predictive.

\begin{wraptable}{R}{0.41\textwidth}
\vspace{-0.3in}
\centering
\caption{Highest-ranked experts by absolute mean standardized weight. Signs indicate the direction of the absence score.}
\small
\setlength{\tabcolsep}{4pt}
\renewcommand{\arraystretch}{1.05}
\begin{tabular*}{\linewidth}{@{\extracolsep{\fill}}rlr@{}}
\hline
\textbf{Rank} & \textbf{Expert} & \textbf{Mean weight} \\
\hline
1 & L04.E067 & $-0.2675$ \\
2 & L02.E013 & $+0.2274$ \\
3 & L45.E063 & $-0.2225$ \\
4 & L02.E055 & $+0.1949$ \\
5 & L13.E018 & $+0.1821$ \\
6 & L02.E089 & $-0.1817$ \\
7 & L47.E071 & $+0.1793$ \\
8 & L02.E072 & $+0.1778$ \\
\hline
\end{tabular*}
\label{tab:expert-weight-top}
\end{wraptable}

\subsubsection{Expert Weight Distribution}

Table~\ref{tab:expert-weight-top} lists the eight experts with the largest absolute mean standardized coefficients in the full 6,144-feature linear probe, averaged across three training seeds. Positive weights increase the predicted absence score, while negative weights decrease it. Four of these experts are in Layer~2, although experts from later layers also appear in the list. This concentration suggests that target-absence information emerges in an early layer, but the ranking alone cannot establish whether that layer supports detection on its own. Expert weight ranking details are provided in Appendix~\ref{app:expert-weight-ranking}. This early signal is consistent with how models integrate visual information: a vision encoder first extracts image features, which enter the language decoder alongside text tokens, allowing attention to incorporate visual context into target-token representations.
Qwen3-VL additionally injects intermediate vision-encoder features into its first three decoder layers through DeepStack. These pathways could explain why target-token routing becomes informative early.

\begin{figure*}[htbp]
    \centering
    \includegraphics[width=0.92\textwidth]{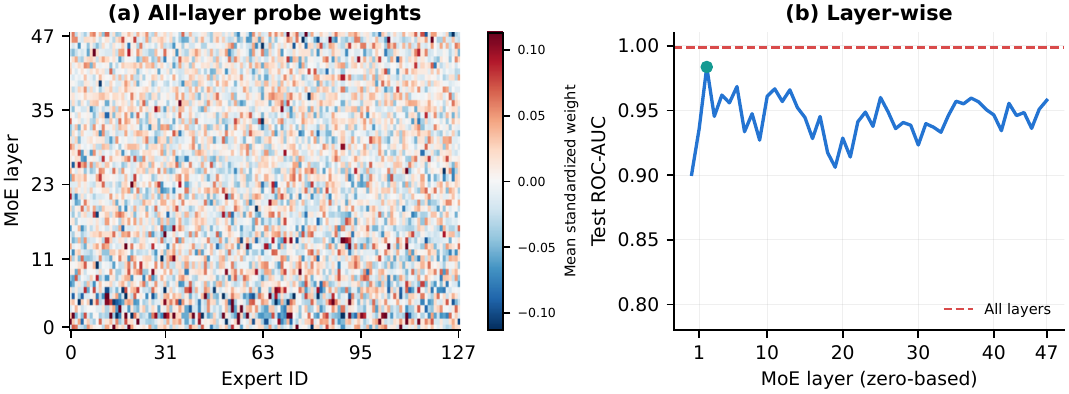}
        \caption{Routing analysis in Qwen. (a) Standardized coefficients of the full 6,144-feature probe, averaged across three training seeds. Red and blue indicate positive and negative weights; the color scale is clipped at the 99th percentile of absolute weights. (b) Test ROC-AUC of 48 single-layer probes. The dashed line shows the full 48-layer probe and the marker highlights Layer~2. 
        % Layer indices start at zero.
        }
        \label{fig:routing-analysis}
\end{figure*}

Figure~\ref{fig:routing-analysis}(a) extends this view to all experts, with one row per layer and one column per expert. Larger positive and negative weights are more visible in the early layers, providing another sign that the absence signal may emerge early. At the same time, weights spread across many layers and experts; the largest coefficient accounts for only 0.164\% of the sum of absolute coefficients. The full probe therefore uses a broad set of features rather than relying on one expert. These weights show how the probe uses routing features, not the causal role of each expert. Feature correlations and regularization also affect weight size, so the heatmap cannot tell us whether an early layer can detect absence by itself. We test this possibility with separate probes for each layer.

\subsubsection{Layer-Wise Detection}
To test the early-layer pattern in the heatmap, we train a separate linear probe on the 128 routing probabilities from each of Qwen's 48 MoE layers. All probes use the same data splits and training procedure, with model selection based on validation performance. Figure~\ref{fig:routing-analysis}(b) compares their test ROC-AUC with the full 6,144-feature probe, shown as a dashed line. Layer~1 is the first to exceed 0.90 validation ROC-AUC, and Layer~2 is the best single layer selected on validation. Layer~2 reaches 0.9836 test ROC-AUC and 94.33\% accuracy using only 128 features, compared with 0.9988 ROC-AUC for the full probe. This confirms that a strong absence signal can already be read from Layer~2, rather than merely receiving large weights in the full probe. Combining all layers still improves detection, so the early-layer probe does not capture all of the useful information. Table~\ref{tab:layerwise-full-results} in the appendix reports test ROC-AUC and accuracy for all 48 classifiers.

%% file: section/6-conclusion.tex
\section{Conclusions}
\label{sec:conclusion}
Target-token routing probabilities in MoE VLMs reveal whether a queried object is absent from the image, even when the model’s answer accepts the false premise.
We use this signal to selectively trigger target-aware visual review, improving final-answer accuracy by up to 22.25 percentage points for Qwen and 13.42 percentage points for Gemma across GQA-Inpaint and OBER, without modifying model weights or router parameters.
Our analyses of Qwen reveal that target-absence information is accessible in early MoE layers and distributed across experts, highlighting a gap between the visual evidence encoded in routing and the model's final answers.
These findings show that routing signals already computed by an MoE VLM can guide selective review and help the model produce answers that better reflect the available visual evidence.

%% file: section/appendix.tex
\section{Appendix}

\subsection{Data and Evaluation Protocol}
\label{app:data-evaluation}

\paragraph{Cohorts and split integrity.}
The probe cohort selects 31,222 GQA-Inpaint image pairs that passed automatic quality control. We use 20,000 pairs for training, 5,446 for validation, and 5,776 for held-out testing. Each pair contributes one target-present and one target-absent single-image example. Images derived from the same source scene remain in the same split, and the probe never receives a pair difference. The GQA review cohort contains 120 pairs, with 60 drawn from the validation split and 60 from the test split. It is used for policy evaluation, not as an independent detector test. The external OBER review cohort contains 115 pairs. These are the primary pairs remaining after 30 of 150 initial candidates were removed during data preparation and five more were excluded for ambiguous image--label semantics.

\begin{table*}[htbp]
\centering
\small
\caption{Cohorts and their roles. Image counts include both present and absent conditions. The 120-pair GQA review cohort includes validation examples; held-out detector performance is measured on the separate 5,776-pair GQA test split.}
\label{tab:cohort-roles}
\begin{tabular}{lrrl}
\hline
Cohort & Pairs & Images & Role \\
\hline
GQA training, automatic QC & 20,000 & 40,000 & Probe fitting \\
GQA validation, automatic QC & 5,446 & 10,892 & Model and threshold selection \\
GQA held-out test, automatic QC & 5,776 & 11,552 & Frozen detector test \\
GQA manually audited review cohort & 120 & 240 & Same-domain policy evaluation \\
OBER manually audited primary cohort & 115 & 230 & External detector and policy evaluation \\
\hline
\end{tabular}
\end{table*}

\paragraph{Manual audit and answer scoring.}
One researcher visually checked that the target is present in each retained original image and that no visually identifiable instance of the same category remains in its edited counterpart. The GQA review cohort also passed automatic quality control. The 120 GQA pairs cover three difficulty groups and four question types, with ten pairs in each difficulty--type cell; half of each cell comes from validation and half from test. All generated answers used in the reported policy comparisons were manually checked against the images. An answer on a target-absent image is marked wrong if it invents the requested attribute, relation, or state; visually supported answers and valid paraphrases are accepted. Five sampled answers from one image are repeated observations, not five independent images.

\subsection{Models, Routing Features, and Probe Training}
\label{app:implementation}

\paragraph{Routing vectors.}
Qwen3-VL-30B-A3B-Instruct has 48 MoE layers with 128 experts per layer, giving 6,144 layer--expert probabilities. Gemma-4-26B-A4B-it has 30 MoE layers with 128 experts per layer, giving 3,840 probabilities. Both models are loaded in bfloat16 and execute the Top-8 selected experts at each layer, but our representation retains the full pre-selection probability distribution. We log routing at the automatically extracted target-object token span during standard prompt prefill and average across sub-tokens for multi-token targets. The two models use separate probes; no model or router weights are changed. On the controlled audited question templates, the extractor recovered the exact target phrase and tokenizer span for all 240 GQA and pre-exclusion OBER questions, including 61 multi-token targets. This result does not establish extraction accuracy on unrestricted questions.

\paragraph{Probe fitting and selection.}
For each layer--expert dimension, standardization uses the GQA training mean and standard deviation only. Each model uses a linear probe trained with binary cross-entropy and AdamW weight decay. Training and selection settings for both models are given in Table~\ref{tab:probe-training-configuration}. We choose weight decay by mean validation ROC-AUC across three training seeds and use the prespecified primary seed for the reported frozen detector. We then select its threshold by validation balanced accuracy. No held-out GQA test or OBER label is used in fitting or selecting these frozen detectors.

\begin{table}[htbp]
\centering
\small
\setlength{\tabcolsep}{4pt}
\caption{Probe training and selection settings. Each model uses the same GQA-Inpaint pair splits (20,000 train, 5,446 validation, and 5,776 test). Standardization and model selection use only training and validation data; thresholds remain frozen on the held-out GQA test and OBER.}
\label{tab:probe-training-configuration}
\begin{tabular}{p{0.38\linewidth}p{0.25\linewidth}p{0.25\linewidth}}
\hline
Setting & Qwen & Gemma \\
\hline
Routing features & 6,144 & 3,840 \\
Feature statistics & Per-feature GQA-train mean/std & Per-feature GQA-train mean/std \\
Loss and optimizer & BCE; AdamW & BCE; AdamW \\
Learning rate; batch size & $10^{-3}$; 512 & $10^{-3}$; 512 \\
Maximum epochs; patience & 30; 5 & 30; 5 \\
Epoch selection & Best validation ROC-AUC & Best validation ROC-AUC \\
Weight-decay candidates & $\{10^{-5},10^{-4},10^{-3}\}$ & $\{10^{-5},10^{-4},10^{-3}\}$ \\
Weight-decay selection & Mean validation ROC-AUC & Mean validation ROC-AUC \\
Selected weight decay & $10^{-4}$ & $10^{-5}$ \\
Training seeds & 20260817--19 & 20260817--19 \\
Primary seed & 20260817 & 20260817 \\
Threshold rule & Validation balanced accuracy & Validation balanced accuracy \\
Frozen threshold $\theta$ & 0.3527936 & 0.4542912 \\
\hline
\end{tabular}
\end{table}

\paragraph{Prompts and decoding.}
Table~\ref{tab:review-prompt-templates} gives the exact review instructions. Each instruction is followed by a blank line, then the unchanged ``Question: \texttt{\{question\}}'' and ``Answer in one short sentence.'' lines, separated by another blank line. The target placeholder is replaced by the extracted phrase $\phi_i$. Qwen separates the two target-aware sentences with a space; Gemma uses a line break. For each image--question--prompt combination, we sample five answers with temperature 0.7, top-$p$ 0.8, top-$k$ 20, and a 64-token output limit.

\begin{table}[htbp]
\centering
\small
\setlength{\tabcolsep}{5pt}
\caption{Prompt instructions prepended to the unchanged question and short-answer request. Standard generation has no added instruction. The generic and target-aware rows reproduce the instruction text used for review; Qwen and Gemma differ only in the separator between the target-aware sentences.}
\label{tab:review-prompt-templates}
\begin{tabular}{p{0.18\linewidth}p{0.73\linewidth}}
\hline
Condition & Added instruction before the question \\
\hline
Standard & None \\
Generic & ``Review the image carefully before answering.'' \\
Target-aware & ``Before answering, carefully verify whether the image actually contains any \texttt{\{target\}}. If it does not, explicitly state that the target is absent.'' \\
\hline
\end{tabular}
\end{table}

\subsection{Additional Detector Results}
\label{app:detector-details}

\paragraph{Frozen thresholds across cohorts.}
Table~\ref{tab:detector-transfer-full} reports the full detector breakdown. Absence recall is the share of target-absent images sent to review; specificity is the share of target-present images left on the standard path. Trigger rate counts all images sent to review. The GQA review cohort is reported separately because it includes validation examples and is not a held-out detector test. On OBER, the AUC decreases for both models. The frozen GQA thresholds also yield different operating points, especially for Qwen, whose OBER absence recall is high but specificity is low.

\begin{table*}[htbp]
\centering
\small
\caption{Detector performance with the GQA-validation-selected threshold frozen for both datasets. Accuracy, recall, specificity, and trigger rate are percentages. GQA audited is a policy cohort that contains validation images.}
\label{tab:detector-transfer-full}
\begin{tabular}{llrrrrr}
\hline
Model & Cohort & ROC-AUC & Accuracy & Absence recall & Specificity & Trigger rate \\
\hline
Qwen & GQA held-out test & 0.9988 & 98.54 & 98.96 & 98.11 & 50.42 \\
Qwen & GQA audited review & 0.9995 & 98.75 & 99.17 & 98.33 & 50.42 \\
Qwen & OBER external & 0.8095 & 65.22 & 99.13 & 31.30 & 83.91 \\
\hline
Gemma & GQA held-out test & 0.9956 & 96.68 & 97.07 & 96.30 & 50.39 \\
Gemma & GQA audited review & 0.9998 & 99.17 & 100.00 & 98.33 & 50.83 \\
Gemma & OBER external & 0.7781 & 70.87 & 73.04 & 68.70 & 52.17 \\
\hline
\end{tabular}
\end{table*}

\paragraph{Representation comparison.}
The Qwen input-embedding and hidden-state baselines use the same GQA examples, source-scene splits, and target-token spans as the routing probe. Each representation has 2,048 dimensions, compared with 6,144 routing dimensions. An L2-regularized linear probe is trained and selected separately for each representation. The text-only input embeddings give 0.5000 test ROC-AUC and 50.00\% accuracy. Mean hidden states give 0.9979 ROC-AUC and 98.06\% accuracy; the last hidden layer gives 0.9980 and 98.16\%. All-layer routing gives 0.9988 and 98.54\%. Table~\ref{tab:hidden-layer-full-results} lists each single-hidden-layer probe. Several hidden layers are also highly predictive, so this does not show that routing is better than every hidden-state representation. Each routing coordinate, however, names a layer and expert. This comparison uses Qwen and GQA, not Gemma or OBER.

\begin{table*}[htbp]
\centering
\small
\caption{Qwen target-token hidden-state probes on the GQA-Inpaint held-out test split. Each probe uses one decoder layer and the same data splits as the routing probe. Regularization and thresholds are selected on validation data. Layer indices start at zero; accuracy is a percentage.}
\label{tab:hidden-layer-full-results}
\begin{tabular}{rrr@{\qquad}rrr}
\hline
Layer & ROC-AUC & Accuracy & Layer & ROC-AUC & Accuracy \\
\hline
0 & 0.9948 & 97.22 & 24 & 0.9956 & 97.00 \\
1 & 0.9980 & 98.24 & 25 & 0.9962 & 97.48 \\
2 & 0.9986 & 98.54 & 26 & 0.9961 & 97.27 \\
3 & 0.9987 & 98.69 & 27 & 0.9955 & 97.05 \\
4 & 0.9987 & 98.61 & 28 & 0.9953 & 96.95 \\
5 & 0.9987 & 98.61 & 29 & 0.9948 & 96.73 \\
6 & 0.9982 & 98.40 & 30 & 0.9947 & 96.74 \\
7 & 0.9978 & 98.14 & 31 & 0.9954 & 96.99 \\
8 & 0.9975 & 97.99 & 32 & 0.9954 & 96.87 \\
9 & 0.9977 & 97.95 & 33 & 0.9958 & 97.21 \\
10 & 0.9978 & 98.15 & 34 & 0.9958 & 97.21 \\
11 & 0.9974 & 98.04 & 35 & 0.9958 & 97.16 \\
12 & 0.9972 & 97.92 & 36 & 0.9960 & 97.15 \\
13 & 0.9977 & 98.19 & 37 & 0.9971 & 97.71 \\
14 & 0.9974 & 97.92 & 38 & 0.9975 & 97.83 \\
15 & 0.9972 & 97.74 & 39 & 0.9970 & 97.65 \\
16 & 0.9966 & 97.52 & 40 & 0.9976 & 98.05 \\
17 & 0.9961 & 97.34 & 41 & 0.9982 & 98.22 \\
18 & 0.9952 & 97.10 & 42 & 0.9981 & 98.03 \\
19 & 0.9954 & 97.17 & 43 & 0.9979 & 98.03 \\
20 & 0.9950 & 96.84 & 44 & 0.9979 & 98.10 \\
21 & 0.9956 & 97.05 & 45 & 0.9976 & 97.99 \\
22 & 0.9955 & 96.88 & 46 & 0.9977 & 97.93 \\
23 & 0.9956 & 96.93 & 47 & 0.9980 & 98.16 \\
\hline
\end{tabular}
\end{table*}

\paragraph{OBER threshold sensitivity.}
The frozen GQA-to-OBER results above require no OBER labels for threshold selection. As a separate diagnostic, we split the 115 OBER pairs into 57 calibration and 58 evaluation pairs, keeping both images from a pair together. We repeat this grouped split 200 times. For each repetition, we choose the threshold on the calibration pairs and evaluate it only on the other pairs. Table~\ref{tab:ober-threshold-sensitivity} gives held-out means. For Qwen, mean OBER balanced accuracy rises from 65.22\% with its frozen GQA threshold to 77.75\% after local calibration; for Gemma, it rises from 70.87\% to 73.19\%. These diagnostic numbers are not the zero-shot policy results, because local calibration uses labeled OBER examples. Threshold changes alter recall, specificity, and trigger rate, but do not improve ROC-AUC on a fixed set of examples. They also do not establish why the OBER ranking AUC is lower.

\begin{figure}[t]
\centering
\includegraphics[width=\columnwidth]{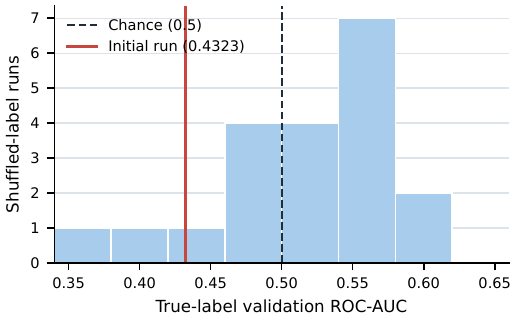}
\caption{Gemma validation ROC-AUC after 20 independently shuffled-label runs. The dashed line marks chance; the red line marks the original single run.}
\label{fig:gemma-permutation-sanity}
\end{figure}

\paragraph{Shuffled-label control.}
An initial Gemma shuffled-label run gave 0.4323 ROC-AUC, outside the original $[0.45,0.55]$ stopping interval. Widening that interval to $[0.4,0.6]$ after observing the result is not independent evidence. We therefore repeated the control 20 times with independently shuffled training and validation labels. Each run selects its epoch with shuffled validation labels and is then evaluated against true validation labels. The mean ROC-AUC is $0.5151\pm0.0618$, consistent with chance; Figure~\ref{fig:gemma-permutation-sanity} shows the full distribution and initial run.

% Newly completed on 2026-09-21 from frozen existing scores/caches.

\begin{table}[t]
\centering
\small
\caption{OBER threshold sensitivity. Calibration uses 200 pair-grouped 50/50 calibration--evaluation splits; the split-calibrated values are held-out means. All rate columns are percentages.}
\label{tab:ober-threshold-sensitivity}
\begin{tabular}{llrrrr}
\hline
Model & Threshold & Balanced acc. & Recall & Specificity & Trigger rate \\
\hline
Qwen & Frozen GQA & 65.22 & 99.13 & 31.30 & 83.91 \\
Qwen & OBER split-calibrated & 77.75 & 94.86 & 60.64 & 67.11 \\
Gemma & Frozen GQA & 70.87 & 73.04 & 68.70 & 52.17 \\
Gemma & OBER split-calibrated & 73.19 & 68.61 & 77.77 & 45.42 \\
\hline
\end{tabular}
\end{table}

\subsection{Additional Review-Policy Results}
\label{app:policy-details}

\paragraph{Full end-to-end accuracy.}
Table~\ref{tab:policy-full} gives the values plotted in the main paper. Each accuracy is computed across both target-present and target-absent images. A reviewed answer replaces the standard answer only when the model-specific detector exceeds its frozen GQA threshold. The target-aware gain is an absolute percentage-point difference from standard answering.

\begin{table*}[htbp]
\centering
\small
\caption{End-to-end answer accuracy for standard generation and two routing-gated review prompts. All values except gain are percentages.}
\label{tab:policy-full}
\begin{tabular}{llrrrr}
\hline
Model & Dataset & Standard & Generic review & Target-aware review & Target-aware gain \\
\hline
Qwen & GQA & 66.75 & 67.67 & 89.00 & +22.25 \\
Qwen & OBER & 83.39 & 89.57 & 95.57 & +12.17 \\
Gemma & GQA & 80.00 & 80.75 & 93.42 & +13.42 \\
Gemma & OBER & 93.91 & 94.09 & 95.30 & +1.39 \\
\hline
\end{tabular}
\end{table*}

\begin{figure}[t]
\centering
\includegraphics[width=\columnwidth]{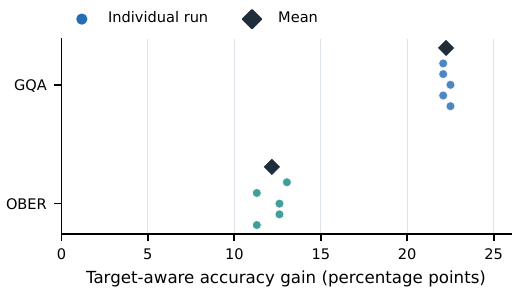}
\caption{Qwen target-aware accuracy gains across five sampled generations per image. Points show individual repetitions; diamonds show dataset means.}
\label{fig:qwen-repeat-stability}
\end{figure}

\paragraph{Repeated Qwen generations.}
Qwen's target-aware gain is positive in all five sampled repetitions on each dataset (Figure~\ref{fig:qwen-repeat-stability}). Mean gains are $22.25\pm0.23$ points on GQA and $12.17\pm0.81$ on OBER. These repetitions test sampling stability on the same images, not independent dataset replication.

\paragraph{Selective versus always review and review-call cost.}
For Qwen, selective and always review use the same 120 GQA and 115 OBER pairs, prompts, and five sampled answers per image. With target-aware review, selective accuracy exceeds always-review accuracy on GQA (89.00\% versus 86.25\%) and OBER (95.57\% versus 95.30\%). The gate sends 50.42\% of GQA images and 83.91\% of OBER images to review, compared with 100\% under always review. Thus, it avoids 49.58\% and 16.09\% of potential review calls, respectively. This comparison evaluates review-call count under both the in-domain and shifted cohorts; it does not measure end-to-end latency, energy, or FLOPs.

\begin{table}[htbp]
\centering
\footnotesize
\caption{Qwen selective versus always review with manually checked answers. Accuracy combines target-present and target-absent inputs. Always review has a 100\% review rate; the final column gives the selective rate.}
\label{tab:selective-always-overall}
\resizebox{\columnwidth}{!}{%
\begin{tabular}{llrrr}
\hline
Dataset & Prompt & Always acc. & Selective acc. & Review rate \\
\hline
GQA & Generic & 67.25\% & \textbf{67.67\%} & 50.42\% \\
GQA & Target-aware & 86.25\% & \textbf{89.00\%} & 50.42\% \\
OBER & Generic & 89.39\% & \textbf{89.57\%} & 83.91\% \\
OBER & Target-aware & 95.30\% & \textbf{95.57\%} & 83.91\% \\
\hline
\end{tabular}%
}
\end{table}

\paragraph{Outcomes by true target presence.}
Table~\ref{tab:presence-strata-full} separates present and absent images. Generic review controls for asking the model to inspect the image without explicitly checking the target. Target-aware review gives much larger gains on absent images but can reduce accuracy on present images, especially when applied to every input.

\begin{table*}[htbp]
\centering
\small
\caption{Qwen accuracy (\%) by true target presence. Each condition has 600 sampled answers on GQA or 575 on OBER. Selective review keeps the standard answer when the gate does not trigger; always review replaces every answer.}
\label{tab:presence-strata-full}
\begin{tabular}{lllrrr}
\hline
Dataset & Target & Review prompt & Standard & Always & Selective \\
\hline
GQA & Absent & Generic & 37.17 & 39.00 & 39.00 \\
GQA & Present & Generic & 96.33 & 95.50 & 96.33 \\
GQA & Absent & Target-aware & 37.17 & 83.17 & 82.50 \\
GQA & Present & Target-aware & 96.33 & 89.33 & 95.50 \\
OBER & Absent & Generic & 66.78 & 79.83 & 79.83 \\
OBER & Present & Generic & 100.00 & 98.96 & 99.30 \\
OBER & Absent & Target-aware & 66.78 & 95.83 & 95.83 \\
OBER & Present & Target-aware & 100.00 & 94.78 & 95.30 \\
\hline
\end{tabular}
\end{table*}

\begin{figure*}[htbp]
\centering
\includegraphics[width=0.87\textwidth]{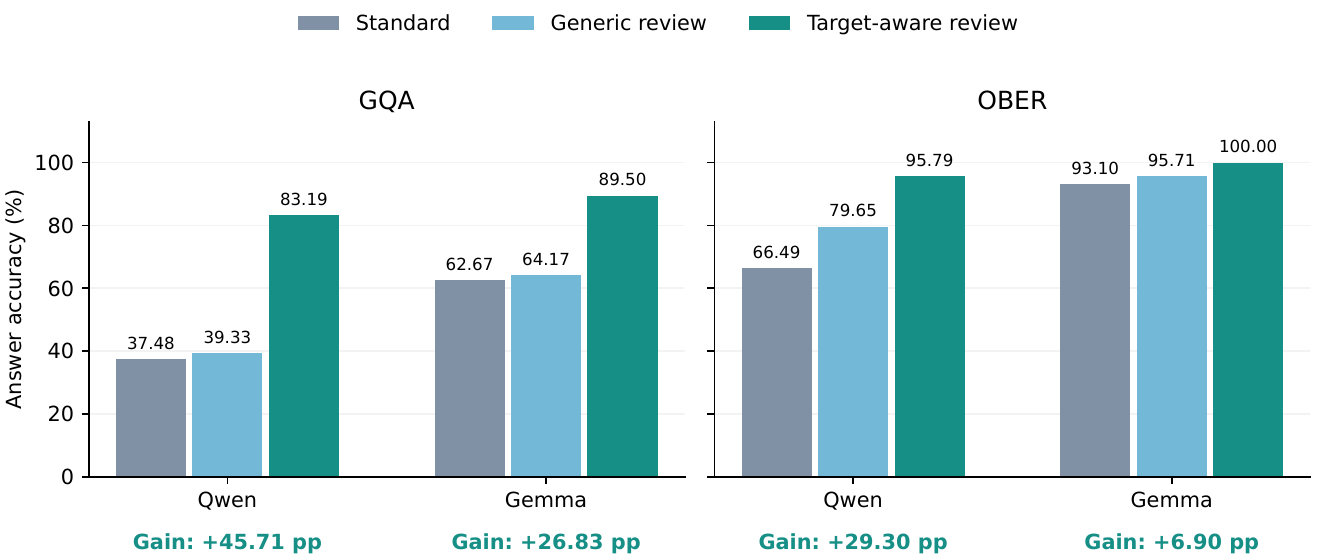}
\caption{Answer accuracy on target-absent images that trigger review, shown separately for GQA and OBER. Within each model and dataset, all three prompts use the same triggered subset with five answers per image. Gains are target-aware accuracy minus standard accuracy, in percentage points.}
\label{fig:triggered-correction}
\end{figure*}

\paragraph{Correction after a true-absence trigger.}
Figure~\ref{fig:triggered-correction} compares standard, generic, and target-aware answers on target-absent images sent to review. This subset excludes absent images missed by the gate, so its denominator differs slightly from the full absent rows of Table~\ref{tab:presence-strata-full}. The triggered subset may also differ between Qwen and Gemma.

Selective and always target-aware review differ only on inputs that the gate predicts are present. On GQA true-present inputs in this group, always review changes 39 correct answers to wrong ones and fixes two wrong answers; on true-absent inputs, it fixes four answers that selective review leaves unchanged. The net selective advantage is therefore 33/1200 answers, or 2.75 points. On OBER, always review changes three correct present answers to wrong ones and fixes none among the skipped inputs; the net advantage is 3/1150 answers, or 0.26 points. A pair-clustered bootstrap (5,000 resamples) gives 95\% intervals of $[1.17,4.58]$ and $[0,0.70]$ points for these overall differences. The OBER gain is small and its interval includes zero.

\paragraph{False-trigger counts.}
Table~\ref{tab:false-trigger-full} expands the false-trigger plot in the main paper. The denominator is the number of initially correct answers on target-present images sent to review, not the number of independent images. Qwen has two such images in GQA and 79 in OBER; Gemma has two and 36. Among Qwen's 405 eligible answers, generic review changes the wording of 314 (77.53\%) but makes only four wrong; target-aware review changes 317 (78.27\%) but makes 32 wrong. A changed answer is therefore not necessarily a harmed answer. Target-aware review changes fewer than 8\% of these initially correct answers to wrong answers after pooling datasets for either model, but the two-image GQA subsets are too small for a stable per-dataset risk estimate. The full policy results above show that this harm does not erase the net accuracy gain.

\begin{table*}[htbp]
\centering
\small
\caption{Correct-to-wrong changes among initially correct answers on false-triggered target-present images. Each cell shows the number of harmed answers over the eligible answer count. Five answers from one image are correlated.}
\label{tab:false-trigger-full}
\begin{tabular}{llrr}
\hline
Model & Dataset & Generic review & Target-aware review \\
\hline
Qwen & GQA & 0/10 (0.00\%) & 5/10 (50.00\%) \\
Qwen & OBER & 4/395 (1.01\%) & 27/395 (6.84\%) \\
Qwen & Combined & 4/405 (0.99\%) & 32/405 (7.90\%) \\
Gemma & GQA & 0/10 (0.00\%) & 0/10 (0.00\%) \\
Gemma & OBER & 10/178 (5.62\%) & 15/178 (8.43\%) \\
Gemma & Combined & 10/188 (5.32\%) & 15/188 (7.98\%) \\
\hline
\end{tabular}
\end{table*}

\subsection{Layer-Wise Routing}
\label{app:layerwise-results}

\paragraph{Single-layer Qwen probes.}
For each of Qwen's 48 MoE layers, we train a separate linear probe using only that layer's 128 routing probabilities. All runs follow the same split and validation-selection rules. Table~\ref{tab:layerwise-full-results} provides test ROC-AUC and accuracy for every layer. Indices are zero-based. Layer~0 reaches 0.8990 validation and 0.9006 test ROC-AUC, so Layer~1 is the first above 0.90 on validation. Layer~2 is best on validation and reaches 0.9836 test ROC-AUC and 94.33\% accuracy, versus 0.9988 ROC-AUC for all layers.

\begin{table*}[htbp]
\centering
\small
\caption{Test results for all 48 Qwen layer-wise classifiers. Each classifier uses 128 routing probabilities from one layer. Model selection and thresholds use validation data only. Accuracy is reported as a percentage.}
\label{tab:layerwise-full-results}
\begin{tabular}{rrr@{\qquad}rrr}
\hline
Layer & ROC-AUC & Accuracy (\%) & Layer & ROC-AUC & Accuracy (\%) \\
\hline

0 & 0.9006 & 83.19 & 24 & 0.9377 & 86.50 \\
1 & 0.9360 & 87.50 & 25 & 0.9599 & 89.34 \\
2 & 0.9836 & 94.33 & 26 & 0.9489 & 87.69 \\
3 & 0.9454 & 89.08 & 27 & 0.9359 & 86.28 \\
4 & 0.9618 & 90.74 & 28 & 0.9407 & 86.92 \\
5 & 0.9559 & 89.58 & 29 & 0.9385 & 86.42 \\
6 & 0.9684 & 91.38 & 30 & 0.9232 & 84.76 \\
7 & 0.9335 & 86.08 & 31 & 0.9398 & 86.44 \\
8 & 0.9474 & 87.92 & 32 & 0.9371 & 86.07 \\
9 & 0.9270 & 84.96 & 33 & 0.9331 & 85.80 \\
10 & 0.9609 & 89.76 & 34 & 0.9463 & 87.78 \\
11 & 0.9667 & 90.61 & 35 & 0.9571 & 89.02 \\
12 & 0.9569 & 88.89 & 36 & 0.9551 & 88.90 \\
13 & 0.9659 & 90.20 & 37 & 0.9595 & 89.36 \\
14 & 0.9522 & 88.89 & 38 & 0.9566 & 89.09 \\
15 & 0.9445 & 87.45 & 39 & 0.9506 & 88.04 \\
16 & 0.9283 & 85.00 & 40 & 0.9465 & 87.81 \\
17 & 0.9453 & 87.25 & 41 & 0.9343 & 86.05 \\
18 & 0.9173 & 84.50 & 42 & 0.9555 & 88.96 \\
19 & 0.9061 & 82.62 & 43 & 0.9460 & 87.44 \\
20 & 0.9286 & 84.88 & 44 & 0.9484 & 87.95 \\
21 & 0.9140 & 83.32 & 45 & 0.9361 & 86.23 \\
22 & 0.9413 & 86.89 & 46 & 0.9509 & 88.20 \\
23 & 0.9487 & 87.29 & 47 & 0.9578 & 89.60 \\
\hline
\end{tabular}
\end{table*}

\subsection{Top-K Routing Probes}
\label{app:topk-probes}

We rank Qwen's 6,144 routing dimensions by absolute standardized weight in the full linear probe, using training data only. For each prespecified $K\in\{10,50,100,500,1000,1500,2048,4000,6144\}$, we fit a new linear probe on the top $K$ dimensions. All probes use the same training split, optimizer settings, and validation-based epoch and threshold selection. Figure~\ref{fig:topk-probes} shows the held-out test results in the main paper. This experiment does not measure VLM latency or FLOPs, because the router still computes its full probability distribution.

\subsection{Expert Weight Ranking Details}
\label{app:expert-weight-ranking}

All 6,144 Qwen layer--expert coordinates are ranked by the absolute value of their mean standardized coefficient over three training seeds. Positive coefficients increase the predicted absence score, while negative coefficients decrease it. The top eight appear in Table~\ref{tab:expert-weight-top}; Figure~\ref{fig:routing-analysis}(a) shows the weight distribution across all layers. The complete numeric ranking is supplied as the machine-readable file \texttt{all\_6144\_experts\_linear\_weight\_ranking.csv} in the supplementary materials rather than repeated as a long PDF table. This is a ranking within the joint linear probe, not standalone expert accuracy or causal importance.